%% file: main.tex
\let\mypdfximage\pdfximage
\def\pdfximage{\immediate\mypdfximage}
\documentclass[10pt,twocolumn,letterpaper]{article}

\input{preamble}

\input{definitions}

\input{figures.tex}

\input{tables.tex}
\usepackage[pagebackref=true,breaklinks=true,letterpaper=true,colorlinks,bookmarks=false]{hyperref}

\iccvfinalcopy %

\def\iccvPaperID{12403} %

\ificcvfinal\fi

\begin{document}

\title{Unsupervised Learning of Object-Centric Embeddings \\ for Cell Instance Segmentation in Microscopy Images}

\author{
	Steffen Wolf$^{1}$, Manan Lalit$^{2}$, Henry Westmacott$^{1}$, Katie McDole$^{1}$\thanks{Corresponding Author, kmcdole@mrc-lmb.cam.ac.uk.}, Jan Funke$^{2}$\thanks{Corresponding Author, funkej@janelia.hhmi.org.} \\
	\normalsize{$^1$MRC Laboratory of Molecular Biology}, \normalsize{$^2$HHMI Janelia Research Campus}\\
}

\maketitle
\ificcvfinal\thispagestyle{empty}\fi

\input{chapters/00_abstract.tex}
\input{chapters/01_introduction}

\input{chapters/02_related_work.tex}
\input{chapters/03_methodology.tex}

\input{chapters/04_experiments.tex}

\input{chapters/05_discussion.tex}

{\small
\bibliographystyle{ieee_fullname}
\bibliography{references}
}
\clearpage
\onecolumn
\begin{appendices}
\input{chapters/06a_appendix}

\newpage
\input{chapters/06b_appendix}
\newpage
\input{chapters/06c_appendix}
\newpage
\input{chapters/06f_appendix}
\end{appendices}
\end{document}

%% file: definitions.tex
\def\imgspace{\ensuremath{\Omega}\xspace}

\def\positioni{\ensuremath{i}\xspace}
\def\positionj{\ensuremath{j}\xspace}
\def\positionsof#1{\ensuremath{\imgspace_{#1}}\xspace}
\def\positionsofsame#1#2{\ensuremath{\imgspace_{#1}^{#2}}\xspace}
\def\positionsofdifferent#1#2{\ensuremath{\overline{\imgspace}_{#1}^{#2}}\xspace}

\def\patcha{\ensuremath{a}\xspace}
\def\patchb{\ensuremath{b}\xspace}
\def\patchat#1{\ensuremath{p(#1)}\xspace}
\def\patchspace{\ensuremath{\mathcal{P}}\xspace}

\def\emb{\ensuremath{r}\xspace}
\def\embat#1{\ensuremath{\emb(#1)}\xspace}
\def\embparameters{\ensuremath{\theta}\xspace}
\def\embfunc{\ensuremath{f_\embparameters}\xspace}

\def\intraobjectoffset{\ensuremath{\overrightarrow{\positioni\positionj}}\xspace}
\def\invintraobjectoffset{\ensuremath{\overrightarrow{\positionj\positioni}}\xspace}
\def\imgdist{\ensuremath{d}\xspace}
\def\embdist{\ensuremath{\hat{d}}\xspace}
\def\distmeasure{\ensuremath{\sigma}\xspace}
\def\sampleradius{\ensuremath{\kappa}\xspace}

\def\allpairs{\ensuremath{P}\xspace}

\def\centroid{\ensuremath{e}\xspace}

\def\lossoce{\ensuremath{\mathcal{L}_\text{\textsc{OCE}}}\xspace}
\def\losscellulus{\ensuremath{\mathcal{L}}\xspace}

\def\expect#1{\mathbb{E}\left[{#1}\right]}

\def\fluohela{\textsc{FluoHela}\xspace}
\def\tissuenet{\textsc{TissueNet}\xspace}
\def\hsc{\textsc{HSC}\xspace}
\def\huseven{\textsc{HU7}\xspace}
\def\PhCPSC{\textsc{PSC}\xspace}
\def\simds{\textsc{Simulated}\xspace}
\def\immuneds{\textsc{Immune}\xspace}
\def\skinds{\textsc{Skin}\xspace}
\def\lungds{\textsc{Lung}\xspace}
\def\pancreasds{\textsc{Pancreas}\xspace}

\newcommand{\new}[1]{#1}

\def\figref#1{Figure~\ref{#1}}

\def\appref#1{Appendix~\ref{#1}}
\def\tabref#1{Table~\ref{#1}}
\def\eqref#1{Equation~\ref{#1}}

\def\eg{\emph{e.g.}\xspace}
\def\ie{\emph{i.e.}\xspace}

\def\partitle#1{\pagebreak[2]\noindent\textbf{#1}}
\usetikzlibrary{spy}
\usetikzlibrary{decorations,decorations.markings,decorations.text}
\usetikzlibrary{math}

\newlength{\charcelldiameter}
\newlength{\charpatchsize}
\tikzstyle{charcell}=[%
  circle,%
  draw=gray,%
  fill=gray,%
  minimum width=\charcelldiameter,%
  inner sep=0]%
\tikzstyle{charpatch}=[%
  rectangle,%
  minimum width=\charpatchsize,%
  minimum height=\charpatchsize,%
  inner sep=0,%
  thick]%
\def\chardrawpatcha#1#2{%
  \node[charpatch,draw=color_patch_a,fill=gray!20!white] at (#1, #2) {%
    \tikz{%
      \clip (0,0) rectangle (\charpatchsize,\charpatchsize);%
      \node[charcell] at (0.5\charcelldiameter+0.5\charpatchsize, 0.5\charpatchsize) {};%
    }%
  };%
}%
\def\chardrawpatchb#1#2{%
  \node[charpatch,draw=color_patch_b,fill=gray!20!white] at (#1, #2) {%
    \tikz{%
      \clip (0,0) rectangle (\charpatchsize,\charpatchsize);%
      \node[charcell] at (0.5\charpatchsize-0.5\charcelldiameter, 0.5\charpatchsize) {};%
    }%
  };%
}%
\def\charpatcha{\raisebox{-0.4mm}{\tikzset{external/export next=false}\tikz{\chardrawpatcha{0}{0}{i}}}\xspace}
\def\charpatchb{\raisebox{-0.4mm}{\tikzset{external/export next=false}\tikz{\chardrawpatchb{0}{0}{j}}}\xspace}
\def\same#1{\emph{\color{patch_j_same!80!black}#1}}
\def\different#1{\emph{\color{patch_j_different!80!black}#1}}

\newcommand{\Cellulus}{\mbox{\textsc{Cellulus}}\xspace}
\newcommand{\StarDist}{\mbox{\textsc{StarDist}}\xspace}
\newcommand{\Cellpose}{\mbox{\textsc{Cellpose}}\xspace}
\newcommand{\EmbedSeg}{\mbox{\textsc{EmbedSeg}}\xspace}
\newcommand{\SEG}{\textsc{SEG}\xspace}
\newcommand{\IoU}{\textsc{IoU}\xspace}
\newcommand{\FOne}{\textsc{F1}\xspace}

\newcommand{\Recall}{\textsc{Recall}\xspace}
\newcommand{\Accuracy}{\textsc{Accuracy}\xspace}
\newcommand{\embedding}{object-centric embedding\xspace}
\newcommand{\embeddings}{object-centric embeddings\xspace}

%% file: figures.tex
\newcommand\figSupervised{
\begin{figure*}[!htb]
\centering
\resizebox{\textwidth}{!}{
\includegraphics{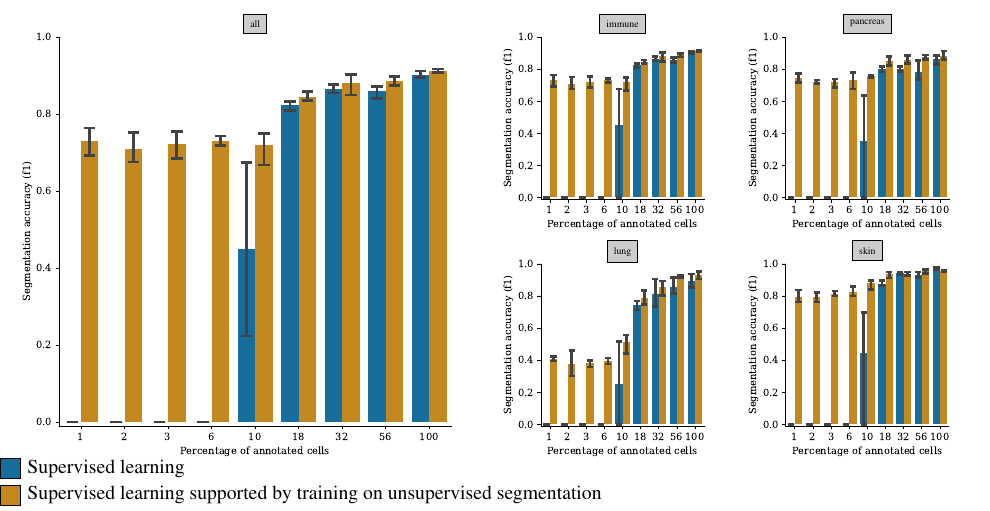}
}
\caption{\textbf{Supervised cell segmentation performance for varying amounts of annotations}. 
We compare a classical supervised learning approach~(blue) trained using only manual annotations, against using a mixture of manual annotations and pseudo-ground truth derived from our unsupervised OCEs (orange), on the four tissue types (\textsc{Immune}, \textsc{Pancreas}, \textsc{Lung} and \textsc{Skin}) in the \textsc{TissueNet} dataset~\cite{greenwald2021whole}. 
The results for \textsc{All} (left) are obtained by averaging the results obtained individually on the four datasets (right).}
\label{fig:supervised}
\end{figure*}
}

%% file: chapters/00_abstract.tex
\begin{abstract}

Segmentation of objects in microscopy images is required for many biomedical applications.
We introduce object-centric embeddings (OCEs), which embed image patches such that the spatial offsets between patches cropped from the same
  object are preserved.
Those learnt embeddings can be used to delineate individual objects and thus
  obtain instance segmentations.
Here, we show theoretically that, under assumptions commonly found in microscopy images, OCEs can be learnt through a self-supervised task that
predicts the spatial offset between image patches.
Together, this forms an unsupervised cell instance segmentation method which we evaluate on nine diverse large-scale microscopy datasets.
Segmentations obtained with our method lead to substantially improved results, compared to state-of-the-art baselines on six out of nine datasets, and perform on par on the remaining three datasets.
If ground-truth annotations are available, our method serves as an excellent starting point for supervised training, reducing the required amount of
ground-truth needed by one order of magnitude, thus substantially increasing the practical applicability of our method.
Source code is available at \url{github.com/funkelab/cellulus}.

\end{abstract}

%% file: chapters/01_introduction.tex
\section{Introduction}
\label{sec:introduction}

Determining whether two image regions belong to the same object is a fundamental challenge in instance segmentation, albeit a simple task for humans. A plausible hypothesis is that humans learn to recognize parts as belonging to a whole by repeatedly observing them in each other's vicinity. We introduce object-centric embeddings (OCEs), which leverage this assumption for unsupervised instance segmentation.
OCEs map image patches in such a way that the spatial offsets between patches cropped from the same object are preserved in embedding space.
We investigate the usage of OCEs in the domain of microscopy imaging and introduce \Cellulus, a method that identifies and segments individual cells in microscopy images.

By relying on reasonable assumptions about microscopy images, namely that~(i) the objects in these
images have a similar appearance and (ii) the objects in
these images are randomly distributed, we show that OCEs can be learnt in an unsupervised fashion.

Cell instance segmentation is crucial for answering important life science questions. 
In recent years, deep learning-based segmentation approaches~\cite{stringer2021cellpose, lalit2021embedding} have achieved the best performance on standard benchmarking datasets, but these approaches rely on large amounts of annotated training data.
Our proposed unsupervised method \Cellulus, in contrast, circumvents the problem of acquiring these manual annotations.

\begin{figure*}[!htb]
\centering
\resizebox{\textwidth}{!}{%
\input{figures/oce/overview.tikz.tex}}
\caption{\textbf{Method overview and example segmentations on diverse datasets}.
Top row:
An unsupervised learning objective gives rise to \embeddings~(OCEs), such that patches extracted from the same object (green boxes) maintain their relative position to each other. 
Predicted densely, these OCEs allow instance segmentation of cells in microscopy images, by using a post-processing step such as mean-shift clustering. 
Bottom row:
Example raw images and dense OCEs/instance segmentations on four datasets spanning different imaging modalities, cell sizes and shapes.
}
\label{fig:intro:overview}
\end{figure*}
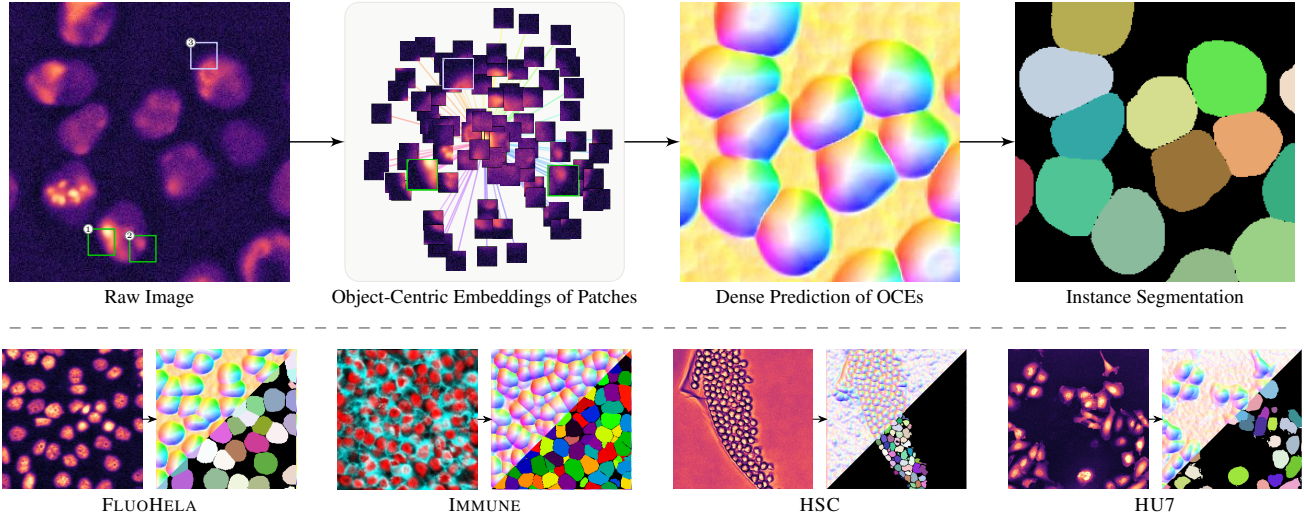

With \Cellulus, we provide an approach for employing the learnt \embedding~locations per patch, identifying image patches that are part of the same cell and thus segmenting cell instances in a unsupervised way (see Figure~\ref{fig:intro:overview} for a few examples).
We demonstrate that this unsupervised segmentation pipeline achieves competitive results with respect to pre-trained baseline models on a diverse set of nine microscopy image datasets (see \tabref{table:unsupervised_segmentation_scores}).

Additionally, instance segmentations obtained through our proposed unsupervised pipeline are excellent starting points to support supervised training when very little manually generated ground truth annotations are available.
We show that we obtain comparable performance to supervised segmentation methods, after fine-tuning on one order of magnitude less data (see Figure \ref{fig:supervised}).

More generally, supervised training supported by unsupervised segmentation is at least as good as purely supervised learning on all investigated datasets, demonstrating that our method dramatically reduces the amount of ground truth annotations needed, and at times not requiring any. 

Reducing or eliminating the need for manual ground truth is of particular importance to biological research, as new light-microscopy methods are capable of generating terabytes of data in a single experiment. 
Manually annotating even small regions of such datasets can take hundreds or thousands of human hours. 
Thus, there is a tremendous need for self-supervised learning methods to help cope with the vast amount of data generated by modern microscopes.
\Cellulus is available at \url{github.com/funkelab/cellulus}.

%% file: figures/oce/overview.tikz.tex
\includegraphics{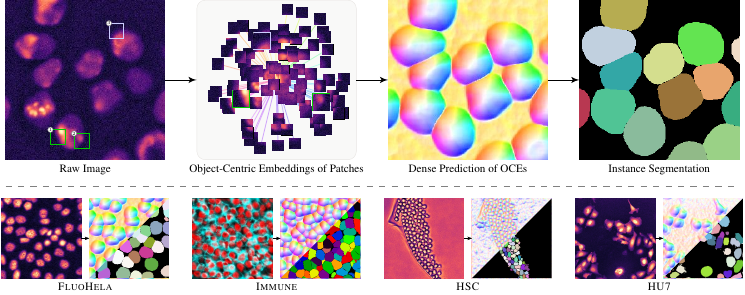}

%% file: chapters/02_related_work.tex
\section{Related Work}\label{sec:related_work}
Currently, machine learning and deep learning-based methods dominate the field of cell instance segmentation~\cite{ulman2017objective, stringer2021cellpose, lalit2021embedding}. 
These cell segmentation methods can be categorized by their intermediate (auxiliary) representation used to derive the predicted segmentation.

\StarDist~\cite{schmidt2018cell}, for example, represents objects as star-convex polygons (\ie, distances from a center point to the cell boundary along sets of equi-distant rays). 
On the other hand, \Cellpose~\cite{stringer2021cellpose} encodes cells by vectors that point inwards from the boundary. 
The representations of \StarDist and \Cellpose are pre-defined and tailored to the tasks of cell segmentation.

Alternatively, pixel-level representations (here referred to as \textit{embeddings}) can be learnt from labels directly by pulling embeddings of pixels within instances together and pushing embeddings across instances apart~\cite{de2017semantic}. 
Initially developed for natural images, this concept was further developed into a cell segmentation and tracking algorithm in the work by Payer \etal~\cite{payer2019segmenting}, which established the state-of-the-art on six Cell Tracking Challenge (CTC) datasets. 

Recent submissions to the CTC further improved the segmentation and tracking performance. 
While Arbelle \etal\cite{arbelle2022dual} and Scherr \etal\cite{scherr2020cell} relied on boundary classification to separate densely clustered cells, Löffler \etal ~\cite{loffler2022embedtrack} used \textit{spatial embeddings}.

Spatial embedding-based approaches learn a function which associates each pixel at location $i$ in the raw image, to a relative spatial embedding (offset vector) $\emb_i$,  such that the resulting absolute spatial embedding $\centroid_i = i \new{+} \emb_i$ for all pixels belonging to an object instance point to a common point (\eg the instance centroid). 

Typically, the embeddings are learnt using a regression loss function, either minimizing the distance between absolute spatial embeddings of pairs of pixels $i,j$ from the same instance
$\mathcal{L}_{\text{regr}} = \sum_{i,j}\distmeasure\left(\centroid_{i}-\centroid_{j}\right) \label{eq:regexample}$
or equivalently by approaching the mean over the whole instance \cite{neven2019instance,lalit2021embedding}. Here, \distmeasure is a measure of distance, \eg, $|\cdot|^2$.
Recently, \EmbedSeg used spatial embeddings to establish the state-of-the-art on multiple 2D and 3D microscopy datasets~\cite{lalit2021embedding}. 

We note that our learning approach has parallels with supervised learning of pixel-wise spatial embeddings.
In our work, self-supervised learning leads to \embeddings, which are post-processed using the mean-shift clustering algorithm, analogous to De Brabandere \etal~\cite{de2017semantic}.

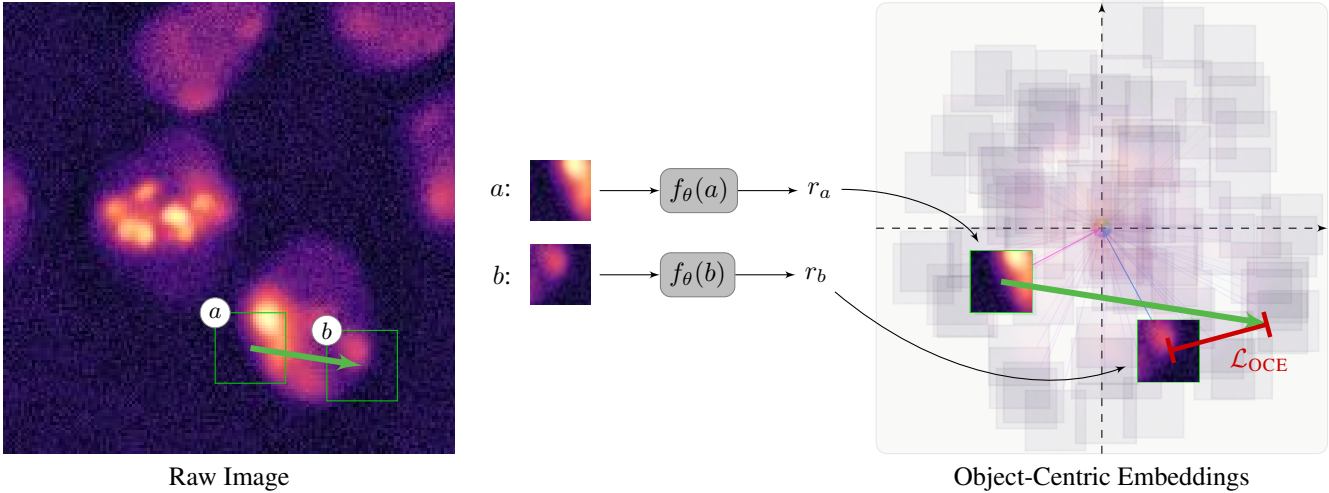
\begin{figure*}[t]\centerline{\input{figures/object_centric_embeddings.tikz}}
\caption{
    \textbf{Unsupervised Learning of Object-Centric Embeddings}.
    During learning, small image patches are randomly cropped from the raw
    image and embedded through a learnable function \embfunc into a 2D
    embedding space. The objective of the loss \lossoce is to ensure that the
    spatial offset between pairs of patches in the raw image (green arrows) is
    preserved in the embedding space (see \eqref{eq:loss_oce}).
  }
  \label{fig:method:oces}
\end{figure*}

Self-supervised learning methods learn representation by solving tasks that predict an intentionally hidden part of the data. 
Predicting the spatial arrangement of image patches provides a rich signal for learning meaningful representations for downstream tasks. 
Spatial tasks include solving jigsaws~\cite{noroozi2016unsupervised}, 
predicting patch rotations~\cite{gidaris2018unsupervised}, or classifying relative patch positions from a grid-like pattern~\cite{doersch2015unsupervised}. 
More recently, contrastive learning between multiple views~\cite{grill2020bootstrap,he2020momentum,tsai2020self,lin2021completer} enabled learning of representations that transfer well to downstream tasks. 
These learnt representations have been shown to reduce the required amounts of annotated data in tasks such as image classification~\cite{henaff2020data} and semantic segmentation~\cite{li2021few,taleb20203d}.

\subsection{Unsupervised Methods for Cell Segmentation}

Recently, methods for cell instance segmentation have been proposed that do not rely on human annotation.

The unsupervised segmentation pipeline proposed by Din and Yu~\cite{din2021training} employed a Convolutional Neural Network (CNN), which when centered on each cell nucleus is tasked to predict a binary mask for each cell. 
The model is trained without any ground-truth and is tasked to predict consistent masks that cover all foreground pixels. 
However, this method still relied on pre-trained networks for locating the nuclei using which the cell segmentations are predicted and can therefore not be considered fully unsupervised. 

Completely unsupervised \textit{instance  separation} has been proposed by Wolf~\etal~\cite{wolf2020instance}, where inpainting networks are used to determine which image regions are independent. 
These independent regions are determined by a hierarchical optimization strategy that continually subdivides the image until all instances are separated. 
In contrast to our proposed method \Cellulus, the post-processing step of Wolf~\etal~\cite{wolf2020instance} is very computationally expensive and does not provide a method for detecting background regions automatically.

Xie \etal~\cite{xie2020} proposed a self-supervised method that employed two proxy tasks of estimating nuclei size and ranking count of nuclei and this enabled the model to mine instance-aware representations from raw data.

%% file: figures/object_centric_embeddings.tikz.tex
\includegraphics{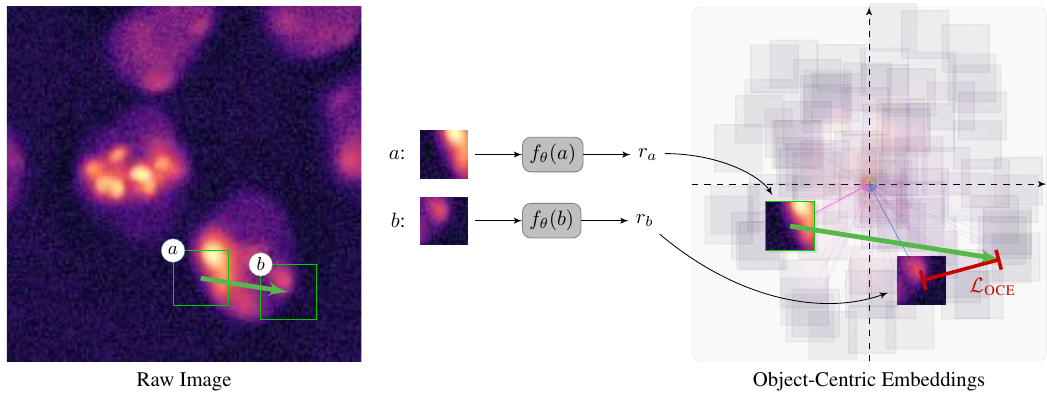}

%% file: chapters/03_methodology.tex
\section{Method}
\label{sec:Method}

We aim to learn an embedding of image patches that reflects the relative spatial arrangement of these patches (\ie the offset between the predicted embeddings  should be equal to their spatial offset), as if they were extracted from the same object ~(see Figure~\ref{fig:method:oces}). 
We refer to the spatial offset between patches extracted from the same object as \textit{intra-object offset} and the learnt embeddings as \emph{object-centric embeddings} (OCEs).

\subsection{Unsupervised Learning of OCEs}
\label{sec:method:unsupervised}

Under conditions that are commonly found in microscopy images, OCEs can be
learnt in an unsupervised manner, \ie, without the provision of segmentation
ground-truth.
Those conditions are:
\begin{enumerate}
\setlength\itemsep{-0.25em}
\item Objects in the image are similar
\item Objects are randomly distributed in the image plane
\item A patch cropped from an object contains enough information to identify
  its position inside the object (\ie, no two parts of an object look exactly
  identical)
\end{enumerate}

Under these conditions, the \emph{expected} offset between two image patches is
proportional to the \emph{intra-object offset} of those two patches, \ie, the
spatial offset between those patches if they were part of the same object.

\begin{figure}[h]
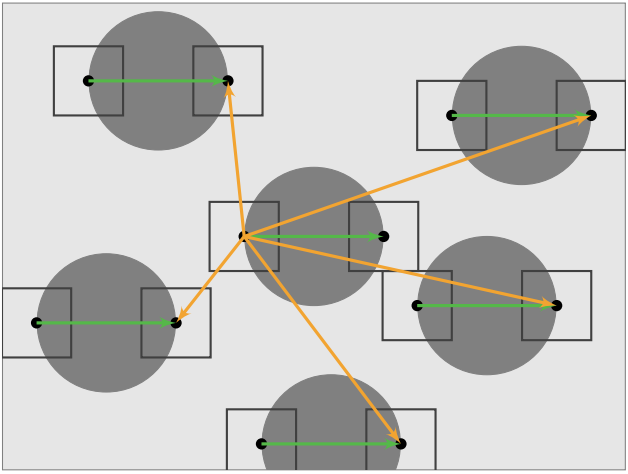

  \include{figures/expected_offset.tikz}
  \vspace{-6mm}
  \caption[]{
    \textbf{Illustration of the expected offset between two example patches in an
    idealized image}:
    Black squares show all image locations where the two patches \charpatcha and
    \charpatchb are found. The expected offset between those patches stems from
    offsets observed within the \same{same} object (\emph{intra-object offsets},
    shown as green arrows) and offsets observed between \different{different}
    objects (\emph{inter-object offsets}, shown as orange arrows for the center
    object only). Assuming a random distribution of objects in large images, the
    average offset between \different{different} objects is zero, thus the
    expected offset \intraobjectoffset between the given patches is proportional
    to the intra-object offset.
  }
  \label{fig:method:expected_offset}
\end{figure}

Let \patcha and \patchb be two different patches found on an object (\eg, the
left- and the right-most patches of a cell, see Figure~\ref{fig:method:expected_offset}).
If multiple similar objects are present in an image, there will be multiple
locations $\positioni \in \imgspace$ where the patch \patcha is visible, and
distinct locations $\positionj \in \imgspace$ where the patch \patchb is
visible.
Here, $\Omega$ is the set of all pixel locations and $x: \Omega \mapsto
\mathcal{R}$ is the image. 

We will refer to the image patch at a location \positioni as
$\patchat{\positioni}$ and denote the set of all locations that contain a given
patch \patcha as \positionsof{\patcha}, \ie, $\positionsof{\patcha} = \left\{
\positioni \in \imgspace \;\;|\;\; \patchat{\positioni} = \patcha \right\}$.

Consider the expected observed offset $\intraobjectoffset = \positionj -
\positioni$ between all occurrences of patches \patcha and \patchb: for each
object contained in the image, patches \patcha and \patchb are observed once
with their \emph{intra-object offset}, \ie, the offset they have to each other
as being part of the same object.
For every pair of different objects, however, patches \patcha and \patchb will
be observed at random offsets, following the assumption that objects in the
image are randomly distributed.
The key insight that allows unsupervised learning of OCEs is that the observed
offsets of patches from different objects have zero mean.

Formally, the expected offset between all locations of two image patches
\patcha and \patchb is given as
\begin{equation}
  \expect{\intraobjectoffset|\patcha,\patchb}
  \approx
  \frac{1}{N}
  \sum_{\positioni \in \positionsof{\patcha}}
    \sum_{\positionj \in \positionsof{\patchb}}
      \intraobjectoffset
  \text{,}
  \label{eq:expectedoffset:allpairs}
\end{equation}
where $N = \left| \positionsof{\patcha} \right| \cdot \left|
\positionsof{\patchb} \right|$ is the number of pairs of image locations $i,j$,
where patches \patcha and \patchb are observed.

This expectation can be rewritten to distinguish observed offsets from the
\emph{same} versus \emph{different} objects. For that, let
\positionsofsame{\patchb}{\positioni} denote all locations \positionj where
patch \patchb appears and is part of the same object at location \positioni.
Similarly, let \positionsofdifferent{\patchb}{\positioni} be the set of
locations \positionj where patch \patchb appears, but is not part of the object
at location \positioni. We can now rewrite the expected observed offset
\intraobjectoffset as
\begin{align}
  \expect{\intraobjectoffset|\patcha,\patchb}
  &\approx
  \frac{1}{N}
  \sum_{\positioni \in \positionsof{\patcha}}
    \left[
      \sum_{\positionj \in \positionsofsame{\patchb}{\positioni}}
        \intraobjectoffset
      +
      \sum_{\positionj \in \positionsofdifferent{\patchb}{\positioni}}
        \intraobjectoffset
    \right]
  \\
  &=
  \underbrace{
    \frac{1}{N_\text{s}}
    \sum_{\positioni \in \positionsof{\patcha}}
      \sum_{\positionj \in \positionsofsame{\patchb}{\patcha}}
        \intraobjectoffset
  }_{\text{intra-object offset}}
  +
  \underbrace{
    \frac{1}{N_\text{d}}
    \sum_{\positioni \in \positionsof{\patcha}}
      \sum_{\positionj \in \positionsofdifferent{\patchb}{\patcha}}
        \intraobjectoffset
  }_{\approx 0}
  \text{,}
  \label{eq:expectedoffset:samedifferent}
\end{align}
where $N_\text{s}$ and $N_\text{d}$  denote the number of times that patches
\patcha and \patchb are observed in the same object and different objects,
respectively.

The first term in \eqref{eq:expectedoffset:samedifferent} is, by definition,
the intra-object offset, \ie, the quantity we aim to infer.
  The second term is the expected offset between patches \patcha and \patchb if
  both are part of different objects. 
  Under the assumption that multiple similar objects are randomly distributed
  in the image, this expectation is zero: observing patch \patcha relative to
  patch \patchb with offset \intraobjectoffset is just as likely as observing
  them at the inverse offset \invintraobjectoffset.
  Without any supervision, the constants $N_\text{s}$ and $N_\text{d}$ are not
  known. The expected offset, calculated as in
  \eqref{eq:expectedoffset:allpairs}, is thus proportional to the sought after
  intra-object offset.

In conclusion, the expected offset given any two patches approximates the
offset between the patches extracted from the same object.
  We can leverage this property to devise a loss function that minimizes the
  differences between the spatial and embedding offsets between pairs of
  patches, and thus learn an object-centric embedding in an unsupervised
  fashion.

Let $\embfunc:\patchspace\mapsto\mathbb{R}^2$ be a parameterized embedding
function, mapping from the set of all image patches \patchspace to a 2D
embedding space. We denote a patch located at \positioni as
\patchat{\positioni} and its embedding as $\embat{\positioni} =
\embfunc(\patchat{\positioni})$.
  We propose the following unsupervised loss, minimizing the difference between
  $\imgdist(\positioni, \positionj) = \positioni - \positionj$ and
  $\embdist(\positioni,\positionj) = \embat{\positioni} - \embat{\positionj}$
  for pairs of patches:
  \begin{equation}
    \lossoce =\sum_{\positioni,\positionj\in\imgspace}
      \distmeasure\left(
        \imgdist(\positioni,\positionj) - \embdist(\positioni,\positionj)
      \right)
    \label{eq:loss_oce}
    \text{,}
  \end{equation}
  where \distmeasure is a measure of distance, \eg, $|\cdot|_2$ (we will
  discuss our choice of \distmeasure below).

\subsection{Loss Implementation}
\label{sec:method:implementation}

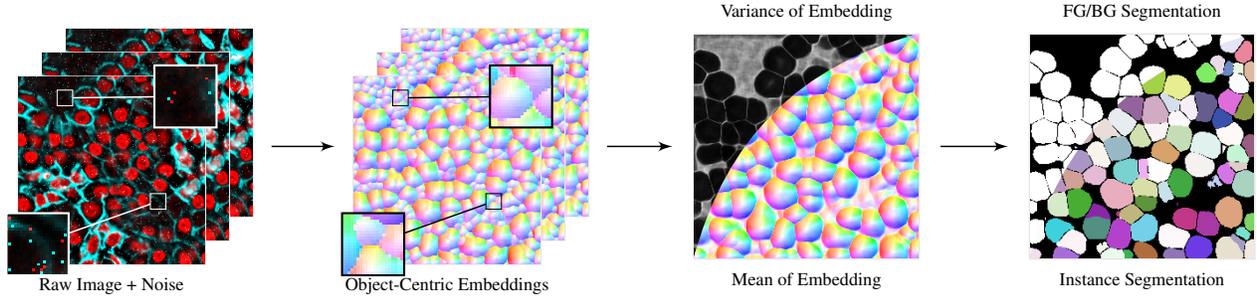
\begin{figure*}[!htb]
\centering
\resizebox{\textwidth}{!}{
\input{figures/pipeline/pipeline}
}
\caption{\textbf{Overview of the inference pipeline}.  
Input image to the trained \embedding (OCE)~network is augmented repeatedly with salt and pepper noise, producing several noisy instances of the raw image (first column). 
OCEs are predicted densely for each noisy instance of the input raw image (second column).
Next, the pixel-wise mean and variance of the predicted OCEs is calculated (third column). 
Images locations with high variance are treated as the background. The remaining foreground region is clustered into individual object instances using mean-shift clustering (fourth column).}
\label{fig:segmentation_pipeline}
\end{figure*}

In practice, the embedding function will be implemented as a Convolutional
Neural Network (CNN) and its weights can be updated using stochastic gradient
descent.
In this setting, strong gradient contributions resulting from pairs of patches of different objects can be problematic due to their high variance, even if they have zero mean.
To address this, we dampen the effect of large distances in our loss function
by using a sigmoid distance function, \ie, $\distmeasure(\delta) =
\left(1+\exp(-\frac{\norm{\delta}^{2}_{2}}{\tau})\right)^{-1}$, where $\tau$ is a hyperparameter controlling the rate of damping.

Furthermore, we limit the sampling of pairs of patches to have a maximal
distance \sampleradius and add an L2 regularization term to obtain our final unsupervised loss function as
\begin{equation}
    \losscellulus =
      \sum_{i,j \in \allpairs}
        \distmeasure\left(
          \imgdist(i,j) - \embdist(i,j)
        \right) +  \lambda_{\text{reg}} \norm{r(i)}_{2}
      \text{,}
      \label{eq:loss_d}
\end{equation} 
where $\allpairs \subset \{\positioni, \positionj \in \imgspace \;|\; |\positioni -
\positionj|_{2} \leq \sampleradius\}$. For more details, see Appendix~\ref{apdx:ssl_training}.

\subsection{Instance Segmentation from OCEs}
\label{sec:method:segmentation}

An instance segmentation can be obtained from OCEs by firstly segmenting
foreground vs.\ background, followed by partitioning the foreground into
individual instances.

To address the background identification, we exploit the sensitivity of the
OCEs to noise in background:
We observe that certain noise patterns in the background (\eg, single bright pixels) become the center point of locally consistent embeddings, thus creating spurious objects (see \figref{fig:segmentation_pipeline}, first column for an example).
To identify background, we repeatedly introduce artificial noise to the raw image and measure the variance of the predicted embeddings (we found salt-and-pepper noise to be effective).
We find that the distribution of the variance of these embeddings over image locations is bi-modal, such that a parameter-free thresholding method like Otsu's is sufficient to separate foreground from background.

After identifying the background, we segment individual instances in the foreground through a mean-shift clustering on the dense OCE predictions~\cite{meanshift,scikit-learn} (see
\figref{fig:segmentation_pipeline}).

%% file: figures/expected_offset.tikz.tex
\includegraphics{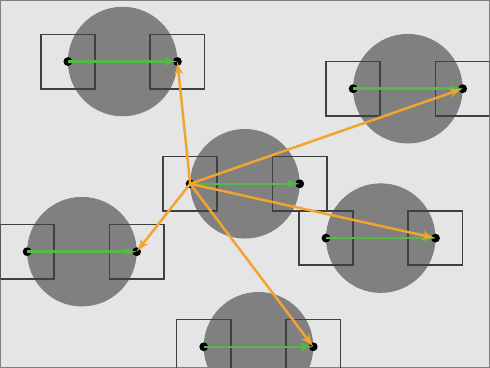}%

%% file: figures/pipeline/pipeline.tex
\newlength{\pipelineimagewidth}%
\setlength{\pipelineimagewidth}{18mm}%
\def\imageoffset{0.6}%

\tikzstyle{image}=[draw=white!80!gray,thick,inner sep=0]
\tikzstyle{closeup} = [
  magnification=4,
  height=\pipelineimagewidth/3,         %
  width=\pipelineimagewidth/3,          %
  connect spies,
  line width=0.80mm
]
\tikzstyle{largewindow}=[line width=0.20mm]
\tikzstyle{arrow}=[->,>=latex']

\includegraphics{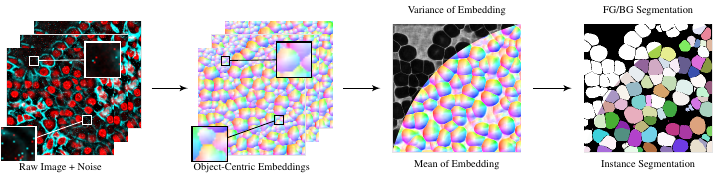}

%% file: chapters/04_experiments.tex
\section{Experiments}\label{sec:Experiments}

\partitle{Used Datasets}.\label{sec:SegmentationDatasets}
We test our method \Cellulus on nine publicly available datasets for which dense ground truth annotations are available. 
The datasets were chosen to represent a diverse set of image modalities, cell/tissue types, and imaging platforms.

\textit{TissueNet}~\cite{greenwald2021whole} is the largest of the analyzed datasets, with $1.3$ million annotated cells. 
It covers six imaging platforms and includes histologically normal and diseased tissue of humans, mice, and macaques. 
The included tissue types (\textsc{Immune}, \textsc{Lung}, \textsc{Pancreas}, \textsc{Skin} cells) vary widely in cell appearance and density. 
Therefore, we add evaluations where we restrict the dataset to the four individual tissue types. 
For reference, constructing \textit{TissueNet} required $>4,000$ hours of human annotation time. 

The nuclei and whole cell are labeled in \textit{TissueNet} and both of these image channels were used during training and inference.
For evaluation purposes during inference, the predicted instance segmentations are compared against the ground truth labels for the whole cell image channel.

\textit{Cell Tracking Challenge (CTC)}~\cite{ulman2017objective} provides diverse 2D and 3D datasets \footnote{\url{http://celltrackingchallenge.net/2d-datasets/}}. 
We select five 2D datasets with distinct cell appearances: \textsc{HSC}, \textsc{HU7}, \textsc{Simulated}, \textsc{FluoHela} and \textsc{PSC}.

Each dataset comes with two sets of image sequences: $1$ and $2$. We used set $1$ for training, while set $2$ is held out for evaluation.
Images in the \textit{CTC} datasets contain only one channel.

\partitle{Segmentation Metrics}. We use two widely used cell segmentation scores:~(i)~\SEG score (used by CTC~\cite{ulman2017objective}) matches every ground truth object to a predicted instance segmentation and measures the average intersection over union (\IoU) of all matches. 
(ii)~\FOne score (used by Greenwald~\etal~\cite{greenwald2021whole}) matches all predictions and ground truth objects with an \IoU greater than or equal to a fixed threshold ($0.5$ unless specified) and reports the \FOne measure of successfully found matches.

\begin{table}[!htb]
\centering
\caption{\textbf{Quantitative results when no annotations are available (\textit{fully unsupervised setting})}. \\
The pretrained models of \StarDist\cite{schmidt2018cell} and \Cellpose\cite{stringer2021cellpose} are compared with \Cellulus on nine diverse microscopy image datasets. Two instance segmentation metrics \FOne and \SEG are evaluated by comparing the quality of predicted instance segmentation with the ground truth instance segmentation. Best performing method on each dataset is shown in bold.
The last row \textsc{TissueNet} (all) shows a weighted average (weights proportional to the number of images) of results for \textsc{Immune}, \textsc{Lung}, \textsc{Pancreas} and \textsc{Skin}.
}
\label{table:unsupervised_segmentation_scores}
\vspace{.7em}
\resizebox{0.5\textwidth}{!}{%
\input{data_tables/ctc_all_hierarchy}
}
\end{table}

\subsection{Unsupervised Segmentation}\label{sec:results:unsupervised}
We compare \Cellulus against two state-of-the-art pretrained segmentation models that are widely used across datasets. 
We investigate the segmentation performance under the condition that no ground truth annotations are available. 

\partitle{Baseline Methods}. 
\StarDist~\cite{schmidt2018cell} is a widely used cell/nucleus segmentation method. 
It predicts, for each pixel, the distances to the boundary in a predefined set of directions. 
\Cellpose~\cite{stringer2021cellpose} uses a supervised network to predict spatial embeddings and clusters pixels together using a diffusion-based aggregation method. 

\partitle{Segmentation Performance. } For each dataset, we train an \embedding~network. 
Raw images are intensity-normalized ($1$ percentile intensity is mapped to $0$ while $99.8$ percentile intensity is mapped to $1$)  and input to the network to produce dense \embeddings. During inference, these \embeddings are processed to obtain instance segmentations and the \FOne and \SEG scores are computed with respect to the ground truth masks for the set of images held-out for evaluation purposes (see \tabref{table:unsupervised_segmentation_scores}).
An overview of the datasets and the predicted embeddings and instance segmentations is shown in Figure~\ref{fig:ctc_extended}.

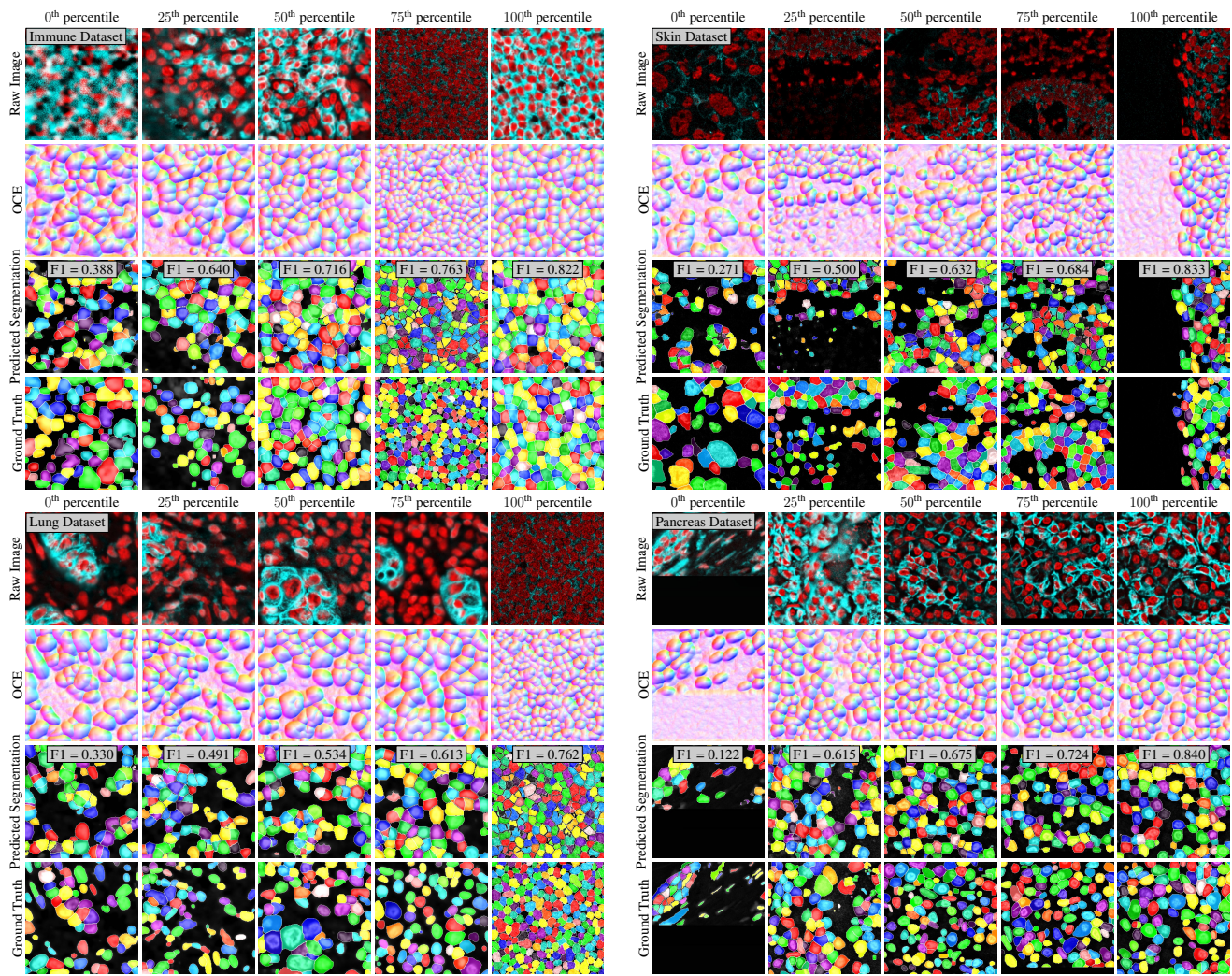
\begin{figure*}[!htb]
\centering
\resizebox{0.49\textwidth}{!}{\input{figures/percentiles/immune_tiles.tikz}}\hfill%
\resizebox{0.49\textwidth}{!}{\input{figures/percentiles/skin_tiles.tikz}}
\resizebox{0.49\textwidth}{!}{\input{figures/percentiles/lung_tiles.tikz}}\hfill%
\resizebox{0.49\textwidth}{!}{\input{figures/percentiles/pancreas_tiles.tikz}}
\caption{\textbf{Predicted OCEs and segmentations on the \tissuenet dataset with tissue types \textsc{Immune} (top-left), \textsc{Skin} (top-right), \textsc{Lung} (bottom-left) and \textsc{Pancreas} (bottom-right)}. 
The \FOne-score is evaluated for each individual image and the $0^{\text{th}}$, $25^{\text{th}}$, $50^{\text{th}}$, $75^{\text{th}}$ and $100^{\text{th}}$ percentile images and their respective \FOne scores are reported in each column.
Rows (from the top) show the Raw Images, Dense Prediction of OCEs, the Predicted Instance Segmentation and the Ground Truth Instance Segmentation available for evaluation purposes.
}
\label{fig:percentiletissuenet}
\end{figure*}

Our method outperforms both baselines on real-world datasets \huseven, \hsc, \simds, \immuneds, \pancreasds and \skinds (according to \SEG score).
On the \simds dataset, our method performs exceptionally well~(see \tabref{table:unsupervised_segmentation_scores} and \appref{apdx:sim_results}). 
To highlight the success and failure modes of our method, we measure the \FOne score per image and report the [$0^{\text{th}}$, $25^{\text{th}}$, $50^{\text{th}}$, $75^{\text{th}}$, and $100^{\text{th}}$] percentile images for different tissue types in \figref{fig:percentiletissuenet}.

\begin{table}[tb]
\centering
\caption{\textbf{ Performance of \textsc{Cellulus} across a range of scale factors for two datasets.}\\
Two instance segmentation metrics \textsc{F1} and \textsc{SEG} are evaluated by comparing the quality of predicted instance segmentation obtained using \textsc{Cellulus} at different scale factors, with the ground truth labels at scale factor = $1.0$. 
Scale factor is inversely related to the employed patch size.
}
\label{tab:resultsPatch}
\vspace{.7em}
\resizebox{\linewidth}{!}{%
\input{data_tables/patch_size}

}
\end{table}

We find that our method can compensate for some variations in object sizes. 
Compare, for example, the small cells in the $75^{\text{th}}$ percentile image of the \immuneds dataset with more voluminous cells in the $100^{\text{th}}$ percentile (see Figure~\ref{fig:percentiletissuenet}). 
However, we also observe that larger outlier objects (\eg, the $0^{\text{th}}$ percentile \skinds image) lead to structural under-segmentation.

\partitle{Background Detection Performance}. We observe that our background detection generally matches the ground truth in the datasets \huseven, \fluohela, \tissuenet, \PhCPSC and \simds, where no additional structure in the background is visible. 
When objects are exceptionally dim, their embeddings may vary with the added noise, which leads to them being treated as background (e.g. see Figure~\ref{fig:percentiletissuenet}, $25^{\text{th}}$ percentile in the \skinds dataset).

The \hsc dataset is exceptionally challenging, with a visible culture plate in the background adding additional structure to the image. 
This leads to additional segments in our method (see Figure~\ref{fig:ctc_extended}). 
All studied methods struggle to predict these segments accurately, with an \FOne score below $0.1$. 
Remarkably, our method receives a high \SEG score compared to the other methods. 
Further analysis of this dataset, including an evaluation of segmentation metrics for all matching thresholds, can be found in \appref{apdx:hsc_results}.

\figSupervised
\partitle{Scale Informs All Parameter Choices}.
\label{sec:scale}The size of the selected image patches determines what object sizes can be detected. 
Patches should be smaller than individual objects but still contain meaningful features. 
To keep all network parameters and training setup constant, we resize the training data. 
Specifically, datasets \huseven, \PhCPSC and \simds are re-scaled by a factor of [$0.5$,$ 2.$, $\frac{2}{3}$], respectively. 
All other datasets were analyzed at their native resolution. 

We also explore the performance of \Cellulus across a range of scale factors for two datasets \immuneds and \lungds (see Table \ref{tab:resultsPatch}).
Note that predictions produced for any scale factor are compared against ground truth labels at scale factor = $1.0$ to obtain the \FOne and \SEG scores.

\partitle{Implementation.}
For learning the \embeddings, we use a U-Net architecture with a limited field of view of $16\times16$ (single $2$x down-sampling layer, ReLU activation). For more details, see \appref{apdx:ssl_training}.

After the training, a scale-appropriate bandwidth of mean-shift clustering has to be chosen. We use the implementation  of mean-shift clustering provided by \textit{scikit-learn}~\cite{scikit-learn} and perform a line search to determine the optimal value.

When instances are tightly packed, our segmentation matches closely with the ground truth without post-processing. 
However, when instances are surrounded by background, we find that patches close to the object borders get mapped to the object center. 
Therefore, we shrink our objects to correct for this halo (see \appref{apdx:sim_results}). 
We pick the optimal shrinkage distance between $0$ and $6$ pixels for all datasets and report the best score.

\subsection{Supporting Supervised Learning}\label{sec:supervised}

In the following experiments, we investigate how our unsupervised segmentation can be used to increase model performance when only a few objects are annotated.

\partitle{Supervised Training Setup.} For the supervised training setup, we build two sparsely annotated supervised datasets, which we call the \textit{sparse} and \textit{pseudo dataset} by randomly sampling ground truth objects as a fixed percentage of annotated cells. (1) The \textit{sparse dataset} contains only the annotated samples. (2) The \textit{pseudo dataset}  uses our predicted segmentations as a starting point (pseudo ground truth) and utilizes the same sampled annotations to correct our predictions. 
We mask all our predicted objects that overlap with the annotations and use the annotations instead. We include annotations of background pixels close to the labeled object ($<30$ pixels).

We use these datasets to train a U-Net using a supervised \StarDist training loss $\mathcal{L}_{\text{\StarDist}}$ \cite{schmidt2018cell}. Mini-batches contain half of the images from the \textit{sparse dataset} and the other half from the \textit{pseudo dataset}. 
The \StarDist loss is computed on each respective half ($\mathcal{L}_{\text{\StarDist}}^{\text{sparse}}$ and $\mathcal{L}_{\text{\StarDist}}^{\text{pseudo}}$). 
The total loss  $\mathcal{L}_{\text{\StarDist}} = (1-\alpha) \mathcal{L}_{\text{\StarDist}}^{\text{sparse}} + \alpha \mathcal{L}_{\text{\StarDist}}^{\text{pseudo}}$
is a linear combination, where $\alpha=0$ corresponds to classical supervised training. For further details, see \appref{apdx:supervised_training}.

We train \StarDist models with varying amounts of annotations and compare the performance of models trained only on annotated images ($\alpha=0$, blue in \figref{fig:supervised}) with those supported by our segmentation ($\alpha=0.5$, orange in \figref{fig:supervised}). 
Each experiment is repeated $3$ times with different annotation samples. 
We evaluate the trained networks on the full \tissuenet dataset as well as subsets of tissue types \textsc{Immune}, \textsc{Pancreas}, \textsc{Lung} and \textsc{Skin}. 
All measured performances and their standard deviations are visualized in \figref{fig:supervised}.

\vspace{1em}
\partitle{Supported supervision makes a significant improvement at $1$\%.} When $1$\% of cell annotations from the \tissuenet dataset are used, \FOne$ =0.75\pm0.03$ is obtained which is significantly better than the performance of the purely unsupervised segmentation at \FOne$=0.64$ (see \tabref{table:unsupervised_segmentation_scores}). 
This effect could be due to the biases of the \StarDist representation, which might help to refine the unsupervised segmentation. 

We additionally perform a training experiment using $0$\% ground truth annotations (\ie $\mathcal{L}_{\text{\StarDist}} =\mathcal{L}_{\text{\StarDist}}^{\text{pseudo}}$) and notice no improvement (\FOne$=0.63$). 
In conclusion, the combination of minimal annotations and the supporting pseudo ground truth significantly help.

\partitle{Supported supervision substantially outperforms purely supervised training.} Our proposed supported supervision method can be used as a replacement for training only on annotations without a performance compromise across all annotation levels. 
Notably, at annotation levels $\le10\%$ our method outperforms the baseline substantially.

%% file: data_tables/ctc_all_hierarchy.tex
\begin{tabular}{lrr|rr|rr}
\toprule
{} & \multicolumn{2}{c}{\Cellpose} & \multicolumn{2}{c}{\StarDist} & \multicolumn{2}{c}{\new{\Cellulus}} \\
{} &       \FOne &   \SEG &       \FOne &   \SEG &    \FOne &   \SEG \\
\midrule
\hsc      &     $0.00$ &  $0.00$ &     $\textbf{0.09}$ &  $0.14$ &  $0.06$ &  $\textbf{0.42}$ \\
\huseven &     $0.40$ &  $0.27$ &     $0.03$ &  $0.02$ &  $\textbf{0.75}$ &  $\textbf{0.55}$ \\
\simds  &     $0.49$ &  $0.34$ &     $0.23$ &  $0.35$ &  $\textbf{0.83}$ &  $\textbf{0.65}$ \\
\fluohela       &     $0.36$ &  $0.65$ &     $\textbf{0.38}$ &  $\textbf{0.79}$ &  $0.34$ &  $0.70$ \\
\PhCPSC & $\textbf{0.76}$ &  $\textbf{0.58}$ &     $0.64$ &  $0.47$ &  $0.64$ &  $0.51$ \\
\midrule
\textsc{Immune} &     $0.44$ &  $0.21$ &     $0.66$ &  $0.41$ &  $\textbf{0.69}$ &  $\textbf{0.57}$ \\
\textsc{Lung}  & $0.76$ &  $0.53$ &     $\textbf{0.81}$ &  $\textbf{0.59}$ &  $0.51$ &  $0.51$ \\
\textsc{Pancreas}  &     $0.56$ &  $0.36$ &     $0.58$ &  $0.36$ &  $\textbf{0.67}$ &  $\textbf{0.49}$ \\
\textsc{Skin} &     $0.48$ &  $0.26$ &     $0.39$ &  $0.24$ &  $\textbf{0.60}$ &  $\textbf{0.46}$ \\
\midrule
\textsc{TissueNet} (all) & 0.55 & 0.32 & 0.59 & 0.38 & \textbf{0.64} & \textbf{0.52} \\
\bottomrule
\end{tabular}

%% file: data_tables/patch_size.tex
\begin{tabular}{crrrrrrrrrrr}
    \toprule
  Scale Factor  & $0.5$ & $0.6$ & $0.7$ & $0.8$ & $0.9$ & $1.0$ & $1.1$ & $1.2$ & $1.3$ & $1.4$ & $1.5$\\ \midrule
     \multicolumn{12}{c}{\textsc{Immune}} \\
     \midrule
       \multicolumn{1}{c}{\textsc{F1}}  & 0.12 & 0.33 & 
       0.48  & 0.44 & 0.64 & 0.69 & 0.69 & 0.66 & 0.67 & 0.59 & 0.58 \\ 
      \multicolumn{1}{c}{\textsc{SEG}}  & 0.20  &
      0.28 & 0.38 & 0.36 & 0.48 & 0.57 & 0.55 & 0.57 & 0.57 & 0.56 & 0.54 \\
    \midrule
    \multicolumn{12}{c}{\textsc{Lung}} \\
    \midrule
    \multicolumn{1}{c}{\textsc{F1}}  & 0.17 &
    0.23 & 0.32 & 0.52 & 0.48 & 0.51 & 0.36 & 0.46 & 0.37 & 0.27 & 0.28 \\  
    \multicolumn{1}{c}{\textsc{SEG}}  & 0.27  & 0.25
      & 0.35 & 0.49 & 0.44 & 0.51 & 0.40 & 0.52 & 0.48  & 0.38 & 0.44\\
    \bottomrule
\end{tabular}

%% file: figures/percentiles/immune_tiles.tikz.tex
\includegraphics{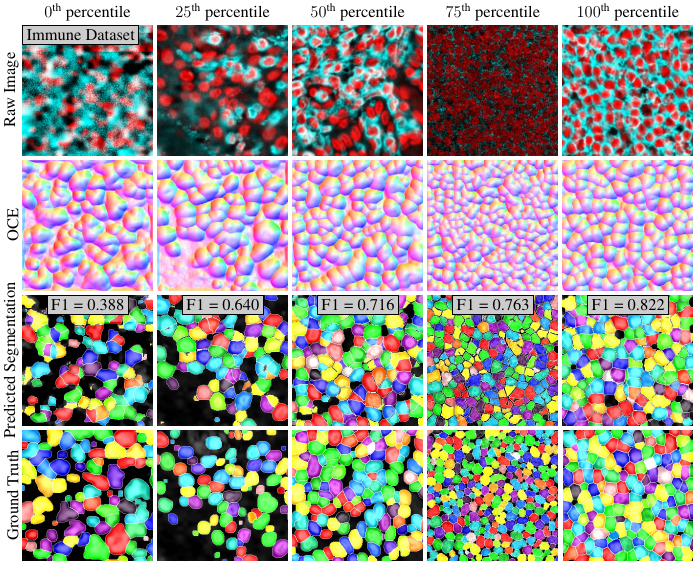}

%% file: figures/percentiles/skin_tiles.tikz.tex
\includegraphics{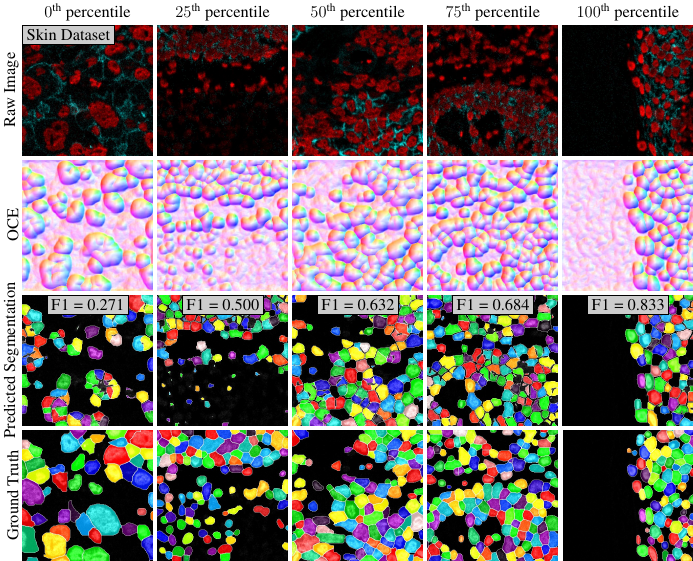}

%% file: figures/percentiles/lung_tiles.tikz.tex
\includegraphics{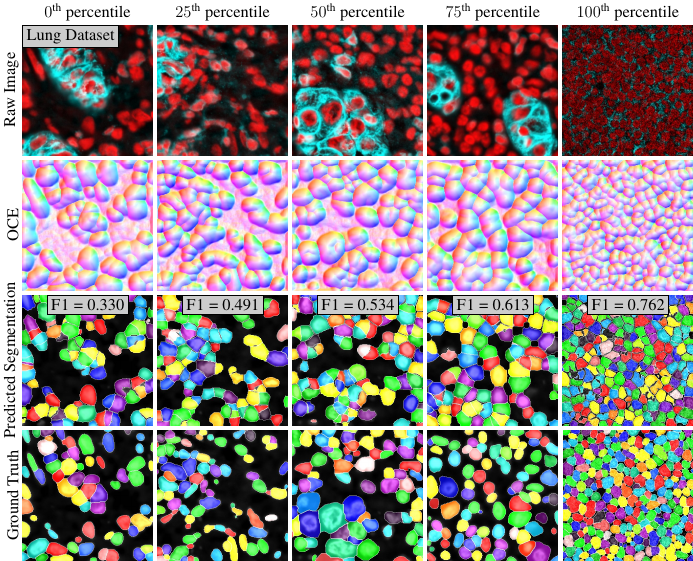}

%% file: figures/percentiles/pancreas_tiles.tikz.tex
\includegraphics{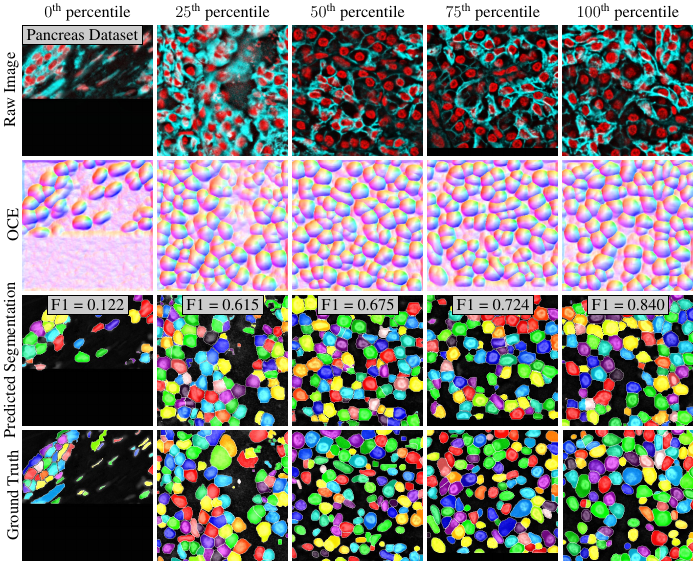}

%% file: chapters/05_discussion.tex
\section{Discussion}

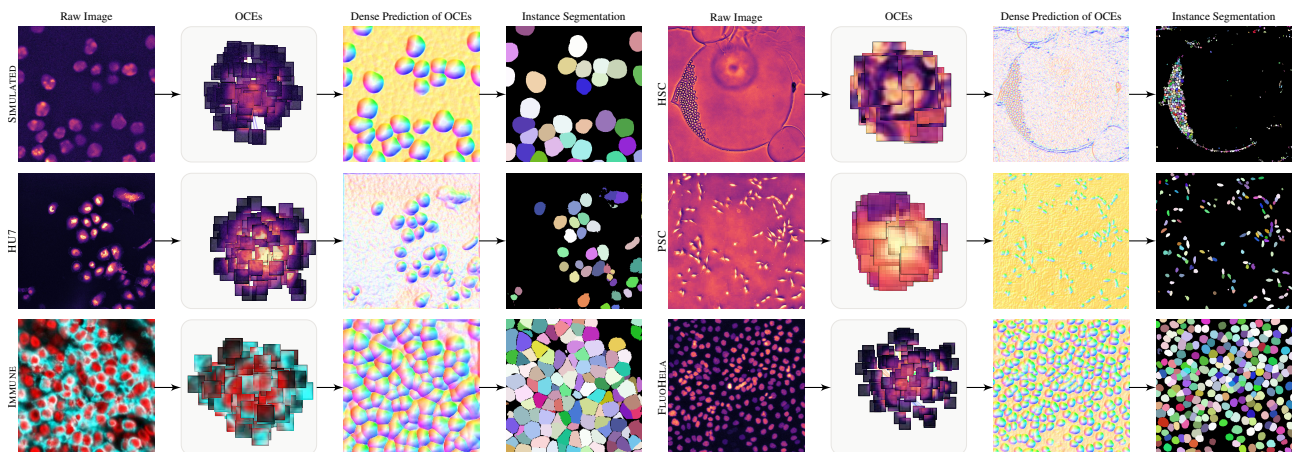
\begin{figure*}[!htb]
\centerline{\input{figures/oce_extended/overview.tikz}}
\caption{\textbf{Qualitative results on a diversity of microscopy image datasets}.
Sample raw images from six datasets (\textsc{Simulated}, \textsc{HU7}, \textsc{Immune}, \textsc{HSC}, \textsc{PSC}, \textsc{FluoHela}) are shown in the first and fifth columns. OCEs are predicted from patches extracted from these raw images and the spatial arrangement of these OCEs are shown in the second and sixth columns. 
During inference, OCEs are predicted densely for the input raw image (third and seventh columns). The final cell instance segmentation can be derived from the dense prediction of OCEs by using a post-processing step such as mean-shift clustering (fourth and eighth columns). }
\label{fig:ctc_extended}
\end{figure*}

We believe that this work offers a feasible way to accelerate the analysis of
microscopy image datasets of cells.
As our experiments on nine large cell segmentation datasets demonstrate (see \tabref{table:unsupervised_segmentation_scores}), a surprisingly good segmentation can often be achieved in a completely unsupervised fashion. 
Depending on the biological question at hand, those results might already be sufficient for downstream analysis.
Furthermore, to obtain more accurate cell segmentations, the segmentations generated in this unsupervised way can be used to augment very small amounts of manual labels and thus increase their efficacy without any additional costs (see  Section~\ref{sec:supervised}). This will in turn drastically reduce the amount of human effort required to analyze large microscopy datasets, and provide a rich source of data for more quantitative and reproducible analyses.  

However, we also note some limitations of our method stemming from violated assumptions:
if objects are not randomly distributed (\eg, if cells always cluster together in pairs), there is no way to tell in a purely unsupervised manner which structure is to be considered as one instance (either the pair of cells, or individual cells).
Similarly, if the objects in the image do not resemble many other examples, the proposed method is unlikely to learn a meaningful \embedding. 
As such, cells with outlier morphologies could result in degenerate segmentations.
Furthermore, we note that the proposed method is sensitive to the size of the
objects to be segmented, \ie, the patch size has to be large enough to contain
enough information to predict the relative position of the patch compared to
others, but small enough to not contain entire objects.
Although this introduces a hyper-parameter that has to be adjusted for each dataset, we believe that this is of little practical relevance since the size of cells in an image can easily be estimated.

While the work discussed here focuses on segmenting cells in 2D datasets, it is theoretically feasible to expand this method to 3D and even 4D datasets. This capability would be particularly useful to biological research, where cells and tissues are commonly imaged in 3D.

{\small
\section*{Acknowledgments}

S.W., H.W. and K.M. are supported by the Medical Research Council, as part of UK Research and Innovation [MCUP1201/23].
}

%% file: figures/oce_extended/overview.tikz.tex
\includegraphics{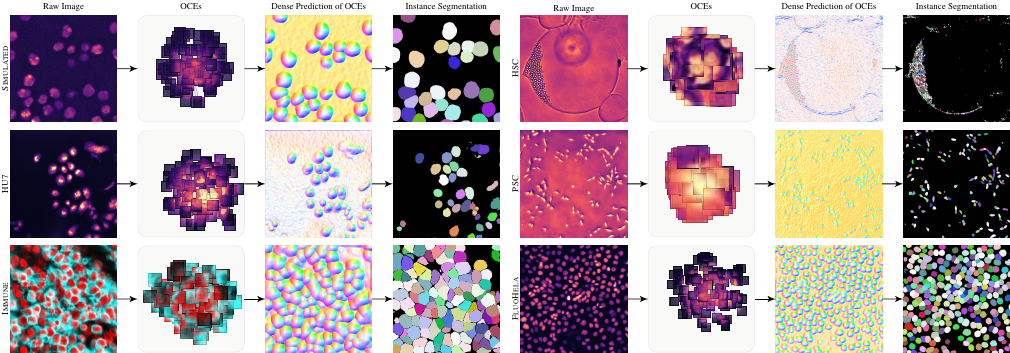}

%% file: chapters/06a_appendix.tex
\section{Self-Supervised Training}\label{apdx:ssl_training}

\partitle{Architecture}. Our self-supervised training requires a network (here referred to as mini U-Net) with a field of view (FoV) smaller than the expected cell diameter.
Since our analyzed datasets contain cells with diameters as small as $20$ pixels wide, we use a U-Net architecture~\cite{ronneberger2015} with an FoV of only $16\times16$. 
To increase the model's capabilities without expanding the FoV, we include additional $1\times1$ convolutions. 
Each U-Net block is composed of a series of [$3\times3$, $1\times1$, $1\times1$, $3\times3$] valid convolution layers with ReLU activations. 
We use a downsampling factor of $2\times2$, a depth of $1$ and constant upsampling layers. 
In the first layer, we use $64$ feature maps and increase it by a factor of $3$ after each block.

\partitle{Training}. We train the mini U-Net on batches of $8$ randomly chosen images with size $252\times252$ pixels. 
We use the Adam optimizer~\cite{kingma2014} with an initial learning rate of $4e^{-5}$ and train for $50$ epochs, reducing the learning rate by a factor of $10$ after epochs $20$ and $30$. 
In our pairwise loss, defined in Equation~\ref{eq:loss_d},

\begin{equation*}
\begin{split}
\mathcal{L} &=
      \sum_{i,j \in \allpairs}
        \distmeasure\left(
          \imgdist(i,j) - \embdist(i,j)
        \right) + \lambda_{\text{reg}} \norm{r(i)}_{2}\text{,} %
\end{split}
\end{equation*}
we use $\distmeasure(\delta) = \left(1+\exp(-\frac{\norm{\delta}^{2}_{2}}{\tau})\right)^{-1}$, $\tau=10$, $\lambda_{\text{reg}}= 1e^{-5}$ and reduce the amount of coordinate pairs to $\allpairs$ to reduce the GPU memory footprint. 
We obtain $\allpairs$ by first sampling $\mathcal{P}_1$ as $10$\% of all pixels. 
For every sample in $p_1 \in \mathcal{P}_1$ we then sample $p_2 \in \mathcal{P}_2$, a random coordinate within radius $\kappa = 10$ of $p_1$. 
In our loss we sample $i,j \in \allpairs = \mathcal{P}_1 \times \mathcal{P}_2$.

%% file: chapters/06b_appendix.tex
\section{Supervised Training}\label{apdx:supervised_training}

\partitle{Architecture.} We use the same architecture as the mini U-Net (see \appref{apdx:ssl_training} ) but increase the depth of the network to $3$ which expands the network's field of view (FoV).

In conclusion, each U-Net block is composed of a series of [$3\times3$, $1\times1$, $1\times1$, $3\times3$] valid convolution layers with ReLU activations. 
We use a downsampling factor of $2\times2$, a depth of $3$ and constant upsampling layers. 
In the first layer, we use $64$ feature maps and increase it by a factor of $3$ after each block.

\partitle{Training}. All models are trained with identical training setups. 
We optimize the loss (see $\mathcal{L}_{\text{\StarDist}}$) with the Adam optimizer with learning rate $1e^{-5}$ for $200$ epochs, reducing the learning rate by a factor of $10$ at epochs $30$, $80$ and $160$. 
We use batches of $8$ images with size $252\times 252$ pixels, sampling pairs of patches within radius $\kappa=10$ (see Equation \ref{eq:loss_d}), and set the loss temperature $\tau=10$.

%% file: chapters/06c_appendix.tex
\section{\hsc Dataset}\label{apdx:hsc_results}
The \hsc dataset is especially challenging - a culture plate is visible in the background, which causes all evaluated models to predict additional object instances near the border of the plate.

We compute scores on a range of \IoU thresholds to investigate robustness of evaluated methods. %
At the matching \IoU threshold of $0.5$, we obtain \FOne scores of $0.00$ for the pre-trained \Cellpose model, $0.09$ for the pre-trained \StarDist model and $0.06$ for \Cellulus \textit{(ours)} (see Table \ref{table:hsc_extra}). 
Additionally for the same matching \IoU threshold, we obtain \Recall scores of $0.01$ for \Cellpose, $0.26$ for \StarDist and $0.55$ for \Cellulus.
On the \hsc dataset, \Cellulus is the most sensitive at detecting cells, but performs less favorably with respect to \FOne and \Accuracy metrics. 
The high \Recall scores obtained with \Cellulus also explains the high \SEG scores where false-positive predictions are not heavily penalized.
Qualitative results on the \hsc dataset can be seen in \figref{fig:ctc_extended}.

\begin{table*}[!htb]
\centering
\definecolor{cellposecolor}{RGB}{216,27,96}
\definecolor{stardistcolor}{RGB}{30,136,229}
\definecolor{ourcolor}{RGB}{255,193,7}
\def\rectcellpose{\tikz{\draw[draw=black,fill=cellposecolor] (11.1,5.5) rectangle ++(0.3,0.3);}}
\def\rectstardist{\tikz{\draw[draw=black,fill=stardistcolor] (11.1,5.5) rectangle ++(0.3,0.3);}}
\def\rectours{\tikz{\draw[draw=black,fill=ourcolor] (11.1,5.5) rectangle ++(0.3,0.3);}}
\caption[]{\textbf{Quantitative results on the \hsc dataset in the Cell Tracking Challenge~\cite{ulman2017objective}, for selected matching \IoU thresholds (\textit{fully unsupervised setting}).} 
Pre-trained \Cellpose and \StarDist baseline models are compared with \Cellulus \textit{(ours)}.
Best performing method on each threshold and metric, is shown in bold. 
}\label{table:hsc_extra}
\resizebox{\textwidth}{!}{%
\input{data_tables/BF-C2DL-HSC}
}
\end{table*}

%% file: data_tables/BF-C2DL-HSC.tex
\begin{tabular}{cccc||ccc||ccc}
\toprule

 & \multicolumn{3}{c}{\Accuracy} & \multicolumn{3}{c}{\FOne} & \multicolumn{3}{c}{\Recall} \\
Threshold & \Cellpose & \StarDist & \Cellulus & \Cellpose & \StarDist & \Cellulus & \Cellpose & \StarDist & \Cellulus \\
\midrule
$0.1$       &     $\textbf{0.68}$ &      $0.22$ &           $0.06$ &      $\textbf{0.81}$ &      $0.36$ &           $0.11$ &      $0.92$ &      $0.98$ &           $\textbf{0.99}$ \\
$0.2$       &      $\textbf{0.59}$ &      $0.21$ &           $0.06$ &      $\textbf{0.74}$ &      $0.35$ &           $0.11$ &      $0.84$ &      $\textbf{0.97}$ &           $0.96$ \\
$0.3$       &      $\textbf{0.27}$ &      $0.18$ &           $0.05$ &      $\textbf{0.43}$ &      $0.30$ &           $0.10$ &      $0.51$ &      $0.83$ &           $\textbf{0.86}$ \\
$0.4$       &      $0.03$ &      $\textbf{0.11}$ &           $0.04$ &      $0.06$ &      $\textbf{0.20}$ &           $0.08$ &      $0.07$ &      $0.55$ &           $\textbf{0.72}$ \\
$0.5$       &      $0.00$ &      $\textbf{0.05}$ &           $0.03$ &      $0.00$ &      $\textbf{0.09}$ &           $0.06$ &      $0.01$ &      $0.26$ &           $\textbf{0.55}$ \\
$0.6$       &      $0.00$ &      $0.01$ &           $\textbf{0.02}$ &     $0.00$ &      $0.03$ &           $\textbf{0.05}$ &      $0.00$ &      $0.07$ &           $\textbf{0.39}$ \\
$0.7$       &      $0.00$ &      $0.00$ &           $\textbf{0.01}$ &      $0.00$ &     $0.00$ &           $\textbf{0.02}$ &      $0.00$ &      $0.01$ &           $\textbf{0.13}$ \\
$0.8$       &      $0.00$ &      $0.00$ &           $0.00$ &      $0.00$ &      $0.00$ &           $0.00$ &      $0.00$ &      $0.00$ &           $\textbf{0.01}$ \\
$0.9$       &      $0.00$ &      $0.00$ &           $0.00$ &      $0.00$ &      $0.00$ &           $0.00$ &      $0.00$ &      $0.00$ &           $0.00$ \\
\bottomrule
\end{tabular}

%% file: chapters/06f_appendix.tex
\section{\simds  Dataset}\label{apdx:sim_results}

\begin{figure*}[!htb]
\centering
\resizebox{\textwidth}{!}{
\input{figures/simsamples/simsamples}}
\caption{\textbf{Qualitative results on the \simds dataset in the Cell Tracking Challenge \cite{ulman2017objective}}. 
The \simds dataset comprises of two time-lapse videos (Videos $01$ and $02$) which contain $64$ and $149$ image frames respectively. 
Shown here are individual raw images from the two videos (first column), predicted instance segmentations obtained using the pre-trained baseline models \StarDist (second column) and \Cellpose (third column), dense prediction of Object-Centric Embeddings (OCEs) obtained using \Cellulus (fourth column), intermediate instance segmentations obtained by applying mean-shift (MS) clustering on the dense OCEs (fifth column), these intermediate instance segmentations are further post-processed (sixth column, see more details in the Section \ref{sec:scale} - \textit{Scale Informs All Parameter Choices}) and the Ground Truth Instance Segmentation available for evaluation purposes (seventh column).\\
Video $02$ in \simds (see last two rows) contains cells with visible granules which cause over-segmentation for both the evaluated, pre-trained baseline models.}
\label{fig:percentiletissuenet_simulated}
\end{figure*}

%% file: figures/simsamples/simsamples.tex
\includegraphics{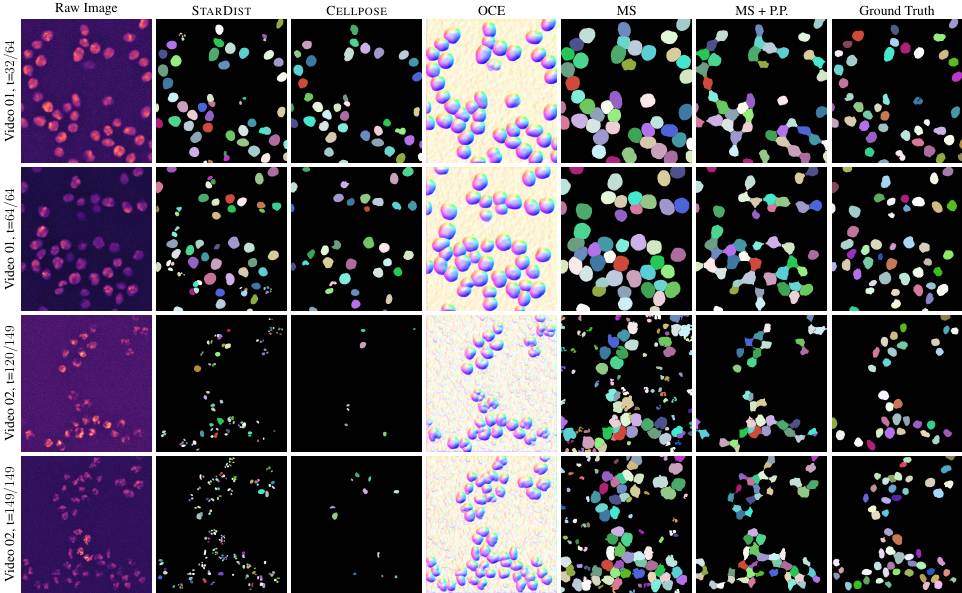}